\documentclass[12pt]{thesis}
\usepackage{graphicx}
\usepackage{authblk}
\usepackage{amsmath}
\usepackage{amssymb}
\usepackage{color}
\usepackage{subcaption}
\usepackage{xparse}
\usepackage{siunitx}

\newcommand{\algo}[1]{\textsc{#1}}
\newcommand{\env}[1]{#1}
\newcommand{\task}[1]{\textsc{#1}}

\DeclareMathOperator{\sgn}{sgn}
\DeclareMathOperator*{\argmax}{arg\,max}

\NewDocumentCommand\E{mg}{%
  \IfNoValueTF{#2}{%
    \rm I\kern-.3em \mathbb{E}\left[#1\right]
  }{%
    \rm I\kern-.3em \mathbb{E}_{#1}\left[#2\right]
  }}

\title{Deep Reinforcement Learning\\[1ex]From Raw Pixels in Doom}
\author{Danijar Hafner}
\university{Hasso Plattner Institute, Potsdam}
\degree{Bachelor of Science}
\degreedate{July 2016}
\supervisor{Prof. Dr. Tobias Friedrich}

\usepackage{tocloft}
\begin{document}

\maketitle

\begin{abstract}
Using current reinforcement learning methods, it has recently become possible
to learn to play unknown 3D games from raw pixels. In this work, we study the
challenges that arise in such complex environments, and summarize current
methods to approach these. We choose a task within the Doom game, that has not
been approached yet. The goal for the agent is to fight enemies in a 3D world
consisting of five rooms. We train the DQN and LSTM-A3C algorithms on this
task. Results show that both algorithms learn sensible policies, but fail to
achieve high scores given the amount of training. We provide insights into the
learned behavior, which can serve as a valuable starting point for further
research in the Doom domain.

\end{abstract}

\begin{romanpages}
\tableofcontents
\end{romanpages}

\chapter{Introduction}
\label{introduction}

Understanding human-like thinking and behavior is one of our biggest
challenges. Scientists approach this problem from diverse disciplines, including
psychology, philosophy, neuroscience, cognitive science, and computer science.
The computer science community tends to model behavior farther away from the
biological example. However, models are commonly evaluated on complex tasks,
proving their effectivity.

Specifically, \emph{reinforcement learning} (RL) and \emph{intelligent
control}, two communities within machine learning and, more generally,
artificial intelligence, focus on finding strategies to behave in unknown
environments. This general setting allows methods to be applied to financial
trading, advertising, robotics, power plant optimization, aircraft design, and
more~\cite{abbeel2007helicopter,vamvoudakis2010aircraft}.

This thesis provides an overview of current state-of-the-art algorithms in RL
and applies a selection of them to the Doom video game, a recent and
challenging testbed in RL.

The structure of this work is as follows: We motivate and introduce the RL
setting in Chapter~\ref{introduction}, and explain the fundamentals of RL
methods in Chapter~\ref{background}. Next, we discuss challenges that arise in
complex environments like Doom in Chapter~\ref{challenges}, describe
state-of-the-art algorithms in Chapter~\ref{algorithms}, and conduct
experiments in Chapter~\ref{experiments}. We close with a conclusion in
Chapter~\ref{conclusions}.

\section{The Reinforcement Learning Setting}

The RL setting defines an environment and an agent that interacts with it. The
environment can be any problem that we aim to solve. For example, it could be a
racing track with a car on it, an advertising network, or the stock market.
Each environment reveals information to the agent, such as a camera image from
the perspective of the car, the profile of a user that we want to display ads
to, or the current stock prices.

The agent uses this information to interact with the environment. In our
examples, the agent might control the steering wheel and accelerator, choose
ads to display, or buy and sell shares. Moreover, the agent receives a reward
signal that depends on the outcome of its actions. The problem of RL is to
learn and choose the best action sequences in an initially unknown environment.
In the field of \emph{control theory}, we also refer to this as a \emph{control
problem} because the agent tries to control the environment using the available
actions.

\section{Human-Like Artificial Intelligence}

RL has been used to model the behavior of humans and artificial agents. Doing
so assumes that humans try to optimize a reward signal. This signal can be
arbitrarily complex and could be learned both during lifetime and through
evolution over the course of generations. For example, he neurotransmitter
\emph{dopamine} is known to play a critical role in motivation and is related
to such a reward system in the human brain.

Modeling human behavior as an RL problem a with complex reward function is not
completely agreed on, however. While any behavior can be modeled as following a
reward function, simpler underlying principles than this could exist. These
principles might be more valuable to model human behavior and build intelligent
agents.

Moreover, current RL algorithms can hardly be compared with human behavior. For
example, a common approach in RL algorithms is called \emph{probability
matching}, where the agent tries to choose actions relative to their
probabilities of reward. Humans tend to prefer the small chance of a high
reward over the highest expected reward. For further details, please refer to
\citet{shteingart2014plausibility}.

\section{Relation to Supervised and Unsupervised Learning}

\emph{Supervised learning} (SL) is the dominant framework in the field of
machine learning. It is fueled by successes in domains like computer vision and
natural language processing, and the recent break-through of deep neural
networks. In SL, we learn from labeled examples that we assume are independent.
The objective is either to classify unseen examples or to predict a scalar
property of them.

In comparison to SL, RL is more general by defining sequential problems. While
not always useful, we could model any SL problem as a one-step RL problem.
Another connection between the two frameworks is that many RL algorithms use SL
internally for function approximation (Sections~\ref{function_approximation}
and~\ref{large_state_space}).

\emph{Unsupervised learning} (UL) is an orthogonal framework to RL, where
examples are unlabeled. Thus, unsupervised learning algorithms make sense from
data by compression, reconstruction, prediction, or other unsupervised
objectives. Especially in complex RL environments, we can employ methods from
unsupervised learning to extract good representations to base decisions on
(Section~\ref{large_state_space}).

\section{Reinforcement Learning in Games}

The RL literature uses different testbeds to evaluate and compare its
algorithms. Traditional work often focuses on simple tasks, such as balancing a
pole on a cart in 2D. However, we want to build agents that can cope with the
additional difficulties that arise in complex environments
(Chapter~\ref{challenges}).

Video games provide a convenient way to evaluate algorithms in complex
environments, because their interface is clearly defined and many games define
a score that we forward to the agent as the reward signal. Board games are also
commonly addressed using RL approaches but are not considered in this work,
because their rules are known in advance.

Most notably, the \env{Atari} environment provided by
\emph{ALE}~\cite{bellemare13ale} consists of 57 low-resolution 2D games. The
agent can learn by observing either screen pixels or the main memory used by
game. 3D environments where the agent observes perspective pixel images include
the driving simulator \env{Torcs}~\cite{torcs}, several similar block-world
games that we refer to as \env{Minecraft} domain, and the first-person shooter
game \env{Doom}~\cite{kempka2016vizdoom} (Section~\ref{doom_domain}).

\chapter{Reinforcement Learning Background}
\label{background}

The field of reinforcement learning provides a general framework that models
the interaction of an agent with an unknown environment. Over multiple time
steps, the agent receives observations of the environment, responds with
actions, and receives rewards. The actions affect the internal environment
state.

For example, at each time step, the agent receives a pixel view of the Doom
environment and chooses one of the available keyboard keys to press. The
environment then advances the game by one time step. If the agent killed an
enemy, we reward it with a positive signal of $1$, and otherwise with $0$.
Then, the environment sends the next frame to the agent so that it can choose
its next action.

\section{Agent and Environment}

Formally, we define the environment as \emph{partially observable Markov
decision process} (POMDP), consisting of a state space $S$, an action space
$A$, and an observation space $X$. Further, we define a transition function
$T\colon S\times A\rightarrow \mathrm{Dist}(S)$ that we also refer to as the
\emph{dynamics}, an observation function $O\colon S\rightarrow
\mathrm{Dist}(X)$, and a reward function $R\colon S\times A\rightarrow
\mathrm{Dist}(\mathbb{R})$, where $\mathrm{Dist}(D)$ refers to the space of
random variables over $D$. We denote $T_{ss'}^a=\Pr(T(s,a)=s')$. The initial
state is $s_0\in S$ and we model terminal states implicitly by having a
recurrent transition probability of~$1$ and a reward of~$0$.

We use $O(s)$ to model that the agent might not be able observe the entire
state of the environment. When $S=X$ and $\forall s\in S\colon\Pr(O(s)=s)=1$,
the environment is fully observable, reducing the POMDP into a \emph{Markov
decision process} (MDP).

The agent defines a policy $\pi\colon\mathcal{P}(X)\rightarrow
\mathrm{Dist}(A)$ that describes how it chooses actions based on previous
observations from the environment. By convention, we denote the variable of the
current action given previous observations as $\pi(x_t)=\pi((x_0,\dots,x_t))$
and the probability of choosing a particular action as
$\pi(a_t|x_t)=\Pr(\pi(x_t)=a_t)$. Initially, the agents provides an action
$a_0$ based on the observation $x_0$ observed from $O(s_0)$.

At each discrete time step $t\in \mathbb{N^{+}}$, the environment observes a
state $s_t$ from $S(s_{t-1},a_{t-1})$. The agent then observes a reward $r_t$
from $R(s_{t-1},a_{t-1})$ and an observation $x_t$ from $O(s_t)$. The
environment then observes $a_t$ from the agent's policy $\pi(x_t)$.

We name a trajectory of all observations starting from $t=0$ an \emph{episode},
and the tuples $(x_{t-1},a_{t-1},r_t,x_t)$ that the agent observes,
\emph{transitions}. Further, we write $E_\pi\left[\cdot\right]$ as the
expectation over the episodes observed under a given policy~$\pi$.

The return $R_t$ is a random variable describing the discounted rewards after
$t$ with discount factor $\gamma\in\left[0,1\right)$. Without subscript, we
assume $t=0$. Note that, although possibly confusing, we stick to the common
notation of using the letter $R$ to denote both the reward function and the
return from $t=0$.

\[
R_t=\sum_{i=1}^\infty{\gamma^i R(s_{t+i},a_{t+i})}.
\]

Note that the return is finite as long as the rewards have finite bounds. When
all episodes of the MDP are finite, we can also allow $\gamma=1$, because the
terminal states only add rewards of $0$ to the return.

The agent's objective is to maximize the expected return $E_\pi\left[R\right]$
under its policy. The solution to this depends on the choice of $\gamma$:
Values close to $1$ encourage long-sighted actions while values close to $0$
encourage short-sighted actions.

\section{Value-Based Methods}
\label{value_based}

RL methods can roughly be separated into \emph{value-based} and
\emph{policy-based} (Section \ref{policy_based}) ones. RL theory often assumes
fully-observable environments, so for now, we assume that the agent found a way
to reconstruct $s_t$ from the observations $x_0,...x_t$. Of course, depending
on $O$, this might not be completely possible. We discuss the challenge of
partial observability later in Section~\ref{partial_observability}.

An essential concept of value-based methods is the value function
$V^\pi(s_t)=\E{\pi}{R_t}$. We use $V^{*}$ to denote the value function under an
optimal policy. The value function has an interesting property, known as
Bellman equation:

\begin{equation}
\label{eq:v_bellman}
V^\pi(s)
=\E{\pi}{R(s,a)+\gamma\sum_{s'\in S}{T_{ss'}^aV^\pi(s')}}.
\end{equation}

Here and in the following sections, we assume $s$ and $a$ to be of the same
time step, $t$, and $s'$ and $a'$ to be of the following time step $t+1$.

Knowing both, $V^{*}$ and the dynamics of the environment, allows us to act
optimally. At each time step, we could greedily choose the action:

\[
\argmax_{a\in A}{\sum_{s'\in S}{T_{ss'}^aV^{*}(s')}}.
\]

While we could try to model the dynamics from observations, we focus on
the prevalent approach of \emph{model-free reinforcement learning} in this
work. Similarly to $V^\pi(s)$, we now introduce the Q-function:

\[
Q^\pi(s,a)=\sum_{s'\in S}{T_{ss'}^aV^\pi(s')}.
\]

It is the the expected return under a policy from the Q-state $(s,a)$, which
means being in state $s$ after having committed to action $a$. Analogously,
$Q^{*}$ is the Q-function under an optimal policy. The Q-function has a similar
property:

\begin{equation}
\label{eq:q_bellman}
Q^\pi(s,a)=R(s,a)+\gamma\E{\pi}{Q^\pi(s',a')}.
\end{equation}

Interestingly, with $Q^{*}$, we do not need to know the dynamics of the
environment in order to act optimally, because $Q^{*}$ includes the weighting
by transition probabilities implicitly. The idea of the algorithms we introduce
next is to approximate $Q^{*}$ and act greedily with respect to it.

\section{Policy Iteration}

In order to approximate $Q^{*}$, the \algo{PolicyIteration} algorithm starts
from any $\pi_0$. In each iteration $k$, we perform two steps: During
\emph{evaluation}, we estimate $Q^{\pi_k}$ from observed interactions, for
example using a Monte-Carlo estimate. We denote this
estimate with $\widehat{Q}^{\pi_k}$. During \emph{improvement}, we update the
policy to act greedily with respect to this estimate, constructing a new
policy~$\pi_{k+1}$ with

\[
\pi_{k+1}(a,s)=\begin{cases}
  1 &\text{if } a=\argmax_{a'\in A}{\widehat{Q}^{\pi_k}(s,a')},\\
  0 &\text{else.}
\end{cases}.
\]

We can break ties in the $\argmax$ arbitrarily. When $\widehat{Q}^{\pi_k}=
Q^{\pi_k}$, it is easy to see that this step is a monotonic improvement,
because $\pi_{k+1}$ is a valid policy and $\forall s\in S,a\in A\colon
\max{a'\in A}Q^{\pi}(s,a')\geq Q^{\pi}(s,a)$. It is even strictly monotonic,
unless $\pi_k$ is already optimal or did not visit all Q-states.

In the case of an estimation error, the update may not be a monotonic
improvement, but the algorithm is known to converge to $Q^{*}$ as the number of
visits of each Q-state approaches infinity~\cite{sutton1998book}.

\section{Exploration versus Exploitation}

In \algo{PolicyIteration}, learning may stagnate early if we do not visit all
Q-states over and over again. In particular, we might never visit some states
just by following $\pi_k$. The problem of visiting new states is called
\emph{exploration} in RL, as opposed to \emph{exploitation}, which means acting
greedily with respect to our current Q-value estimate.

There is a fundamental tradeoff between exploration and exploitation in RL. At
any point, we might either choose to follow the policy that we current think is
best, or perform actions that we think are worse with the potential of
discovering better action sequences. In complex environments, exploration is
one of the biggest challenges, and we discuss advanced approaches in
Section~\ref{exploration}.

The dominant approach to exploration is the straightforward
\algo{EpsilonGreedy} strategy, where the agent picks a random action with
probability $\varepsilon\in \left[0,1\right]$, and the action according to its
normal policy otherwise. We decay $\varepsilon$ exponentially over the course
of training to guarantee convergence.

\section{Temporal Difference Learning}
\label{bootstrapping}

While \algo{PolicyIteration} combined with the \algo{EpsilonGreedy} exploration
strategy finds the optimal policy eventually, the Monte-Carlo estimates have a
comparatively high variance so that we need to observe each Q-state often in
order to converge to the optimal policy. We can improve the data efficiency by
using the idea of \emph{bootstrapping}, where we estimate the Q-values from
from a single transition:

\begin{equation}
\label{eq:bootstrapping}
\widehat{Q}^{\pi}(s_t,a_t)= \begin{cases}
  r_{t+1} &\text{if $s_t$ is terminal,}\\
  r_{t+1}+\gamma\widehat{Q}^{\pi}(s_{t+1},a_{t+1}) &\text{else.}
\end{cases}
\end{equation}

The approximation is of considerably less variance but introduces a bias
because our initial approximated Q-values might be arbitrarily wrong. In
practice, bootstrapping is very common since Monte-Carlo estimates are not
tractable.

Equation~\ref{eq:bootstrapping} allows us to update the estimate
$\widehat{Q}^{\pi}$ after each time step rather than after each episode,
resulting in the online-algorithm \algo{SARSA}~\cite{sutton1998book}. We use a
small learning rate $\alpha\in \mathbb{R}$ to update a running estimate of
$Q^{\pi}(s_t,a_t)$, where $\delta_t$ is known as the \emph{temporal difference
error}:

\begin{equation}
\label{eq:sarsa}
\begin{gathered}
\delta_t=\begin{cases}
  r_{t+1}&-\widehat{Q}^{\pi}(s_t,a_t)
    \quad\text{if $s_t$ is terminal,}\\
  r_{t+1}+\gamma\widehat{Q}^{\pi}(s_{t+1},a_{t+1})&-\widehat{Q}^{\pi}(s_t,a_t)
    \quad\text{else,}
\end{cases}\\[2ex]
\textrm{and }
\widehat{Q}^{\pi}(s_t,a_t)\leftarrow \widehat{Q}^{\pi}(s_t,a_t)+\alpha\delta_t.
\end{gathered}
\end{equation}

\algo{SARSA} approximates expected returns using the current approximation of
the Q-value under its own policy. A common modification to this is known as
\algo{Q-Learning}, as proposed by \citet{watkins1992qlearning}. Here, we
bootstrap using what we think is the best action rather than the action
observed under our policy. Therefore, we directly approximate $Q^{*}$, denoting
the current approximating $\widehat{Q}$:

\begin{equation}
\label{eq:q_learning}
\begin{gathered}
\delta_t=\begin{cases}
  r_{t+1}&-\widehat{Q}(s_t,a_t)
    \quad\text{if $s_t$ is terminal,}\\
  r_{t+1}+\gamma\max_{a'\in A}{\widehat{Q}(s_{t+1},a')})
    &-\widehat{Q}(s_t,a_t)
    \quad\text{else,}
\end{cases}\\[2ex]
\textrm{and }
\widehat{Q}(s_t,a_t)\leftarrow \widehat{Q}(s_t,a_t)+\alpha\delta_t.
\end{gathered}
\end{equation}

The \algo{Q-Learning} algorithm might be one of the more important
breakthroughs in RL \cite{sutton1998book}. It allows us to learn about the
optimal way to behave by observing transitions of an arbitrary policy. The
policy still effects which Q-states we visit and update, but is only needed for
exploration.

\algo{Q-Learning} converges to $Q^{*}$ given continued
exploration~\cite{tsitsiklis1994qoptimal}. This requirement is minimal: Any
optimal method needs to continuously obtain information about the MDP for
convergence.

\section{Eligibility Traces}
\label{eligibility_traces}

One problem, temporal difference methods like \algo{SARSA} and
\algo{Q-Learning} is that updates of the approximated Q-function only affect
the Q-values of predecessor states directly. With long time gaps between good
actions and the corresponding rewards, many updates of the Q-function may be
required for the rewards to propagate backward to the good actions.

This is an instance of the fundamental credit assignment problem in machine
learning. When the agent receives a positive reward, it needs to figure out
what state and action led to the reward so that we can make this one more
likely. The \algo{TD($\lambda$)} algorithm provides an answer to this by
assigning eligibilities to each Q-state.

Upon encountering a reward $r_t$, we apply the temporal difference update rule
for each Q-state $(s,a)$ with an \emph{eligibility trace} $e_t(s,a)\geq0$. The
update of each Q-state uses the received reward scaled by the current
eligibility of the state:

\begin{equation}
\label{eq:eligibility_traces}
\widehat{Q}(s,a)\leftarrow\widehat{Q}(s,a)+\alpha\delta_t e_t(s,a).
\end{equation}

A common way to assign eligibilities is based on the duration between visiting
the Q-states $(s_0,a_0),\ldots,(s_t,a_t)$ and receiving the reward $r_{t+1}$.
The state in which we receive the reward has an eligibility of $1$ and the
eligibility of previous states decays exponentially over time by a factor
$\lambda\in\left[0,1\right]$: $e_t(s_{t-k})=(\gamma\lambda)^k$. We can
implement this by adjusting the eligibilities at each time step:

\[
e_{t+1}(s,a)=\begin{cases}
  1 &\text{if } (s,a)=(s_t,a_t),\\
  \gamma\lambda e_t(s,a) &\text{else.}
\end{cases}
\]

\begin{figure}
\centering
\includegraphics[width=0.62\textwidth]{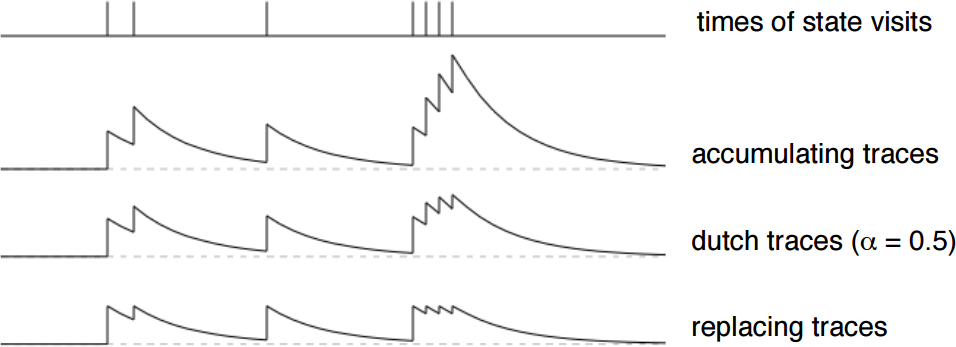}
\caption{Accumulating, Dutch, and replacing eligibility traces
(\citet{sutton1998book}).}
\label{fig:traces}
\end{figure}

This way of assigning eligibilities is known as \emph{replacing traces} because
we reset the eligibility of an encountered state to $1$. Alternatives include
\emph{accumulating traces} and \emph{Dutch traces}, shown in
Figure~\ref{fig:traces}. In both cases, we keep the existing eligibility value
of the visited Q-state and increment it by $1$. For Dutch traces, we
additionally scale down the result of this by a factor between $0$ and $1$.

Using the \algo{SARSA} update rule in Equation~\ref{eq:eligibility_traces}
yields an algorithm known as \algo{TD($\lambda$)}, while using the
\algo{Q-Learning} update rule yields \algo{Q($\lambda$)}.

Eligibility traces bridge the gap between Monte-Carlo methods and temporal
difference methods: With $\lambda=0$, we only consider the current transition,
and with $\lambda=1$, we consider the entire episode.

\subsection{Function Approximation}
\label{function_approximation}

Until now, we did not specify how to represent $\widehat{Q}$. While we could
use a lookup table, in practice, we usually employ a function approximator to
address large state spaces (Section~\ref{large_state_space}).

In literature, gradient-based function approximation is applied commonly. Using
a derivable function approximator like linear regression or neural networks
that start from randomly initialized parameters $\theta_0$, we can perform the
gradient-based update rule:

\begin{equation}
\label{eq:approximate_td}
\theta_{t+1}=\theta_t+\alpha\delta_t \nabla_\theta \widehat{Q}(s_t,a_t).
\end{equation}

Here, $\delta_t$ is the scalar offset of the new estimate from the previous one
given by the temporal difference error of an algorithm like \algo{SARSA} or
\algo{Q-Learning}, and $\alpha$ is a small learning rate. Background in
function approximation using neural networks is out of the scope of this work.

\section{Policy-Based Methods}
\label{policy_based}

In contrast to value-based methods, policy-based methods parameterize the
policy directly. Depending on the problem, finding a good policy can be easier
than approximating the Q-function first. Using a parameterized function
approximator (Section~\ref{function_approximation}), we aim to find a good set
of parameters $\theta$ such that actions sampled from the policy $a\sim
\pi_\theta(a_t,s_t)$ maximize the reward in a POMDP.

Several methods for searching the space of possible policy parameters have been
explored, including random search, evolutionary search~, and gradient-based
search~\cite{deisenroth2013policysurvey}. In the following, we focus on
gradient-based methods.

The \algo{PolicyGradient} algorithm is the most basic gradient-based method
for policy search. The idea is to use the reward signals as objective and tweak
the parameters $\theta_t$ using gradient ascent. For this to work, we are
interested in the gradient of the expected reward under the policy with respect
to its parameters:

\[
\nabla_{\theta}\E{\pi_{\theta}}{R_t}
=\nabla_\theta\sum_{s\in S}{d^{\pi_\theta}(s)
  \sum_{a\in A}{\pi_\theta(a|s)R_t}}
\]
where $d^{\pi_\theta}(s)$ denotes the probability of being in state $s$ when
following $\pi_\theta$. Of course, we cannot find that gradient analytically
because the agent interacts with an unknown, usually non-differentiable,
environment. However, it is possible to obtain an estimate of the gradient
using the \emph{score-function gradient
estimator}~\cite{sutton1999policygradient}:

\begin{equation}
\label{eq:gradient_estimator}
\begin{aligned}
\nabla_{\theta}\E{\pi_{\theta}}{R_t}
&=\nabla_\theta\sum_{s_t\in S}{d^{\pi_\theta}
  \sum_{a_t\in A}{\pi_\theta(a|s)R_t}}\\
&=\sum_{s_t\in S}{d^{\pi_\theta}
  \sum_{a_t\in A}{\nabla_\theta\pi_\theta(a_t|s_t)R_t}}\\
&=\sum_{s_t\in S}{d^{\pi_\theta}\sum_{a_t\in A}{\pi_\theta(a_t|s_t)
  \frac{\nabla_\theta\pi_\theta(a_t|s_t)}{\pi_\theta(a_t|s_t)}R_t}}\\
&=\sum_{s_t\in S}{d^{\pi_\theta}\sum_{a_t\in A}{\pi_\theta(a_t|s_t)
  \nabla_\theta\ln(\pi_\theta(a_t|s_t))R_t}}\\
&=\E{\pi_\theta}{R_t\nabla_\theta\ln\pi_\theta(a_t|s_t)},
\end{aligned}
\end{equation}
where we decompose the expectation into a weighted sum following the definition
of the expectation and move the gradient inside the sum. We then both multiply
and divide the term inside the sum by $\pi_\theta(a|s)$, apply the chain rule
$\nabla_\theta\ln(f(\theta))=\frac{1}{f(\theta)}\nabla_\theta f(\theta)$, and
bring the result back into the form of an expectation.

As shown in Equation~\ref{eq:gradient_estimator}, we only require the gradient
of our policy. Using a derivable function approximator, we can then sample
trajectories from the environment to obtain a Monte-Carlo estimate both over
states and Equation~\ref{eq:gradient_estimator}. This yields the
\algo{Reinforce} algorithm proposed by \citet{williams1992reinforce}.

As for value-based methods, the Monte-Carlo estimate is comparatively slow
because we only update our estimates at the end of each episode. To improve on
that, we can combine this approach with eligibility traces
(Section~\ref{eligibility_traces}). We can also reduce the variance of the
Monte-Carlo estimates as described in the next section.

\section{Actor-Critic Methods}
\label{actor_critic}

In the previous section, we trained the policy along the gradient of the
expected reward, that is equivalent to
$\E{\pi_\theta}{R_t\nabla_\theta\ln\pi_\theta(a|s)}$. When we sample
transitions from the environment to estimate this expectation, the rewards can
have a high variance. Thus, \algo{Reinforce} requires many transitions to
obtain a sufficient estimate.

\emph{Actor-critic methods} improve on the data efficiency of this algorithm by
subtracting a \emph{baseline} $B(s)$ from the reward that reduces the variance
of the expectation. When $B(s)$ is an approximated function, we call its
approximator \emph{critic} and the approximator of the policy function
\emph{actor}. To not introduce bias to the gradient of the reward, the gradient
of the baseline with respect to the policy must
be~$0$~\cite{williams1992reinforce}:

\begin{align*}
\E{\pi_\theta}{\nabla_\theta\ln\pi_\theta(a|s)B(s)}
&=\sum_{s\in S}{d^{\pi_\theta}(s)
  \sum_{a\in A}{\nabla_\theta\pi_\theta(a|s)B(s)}}\\
&=\sum_{s\in S}{d^{\pi_\theta}(s)B(s)\nabla_\theta
  \sum_{a\in A}{\pi_\theta(a|s)}}\\
&=0.
\end{align*}

A common choice is $B(s)\approx V^\pi(s)$. In this case, we train the policy by
the gradient $\E{\pi_\theta}{\nabla_\theta\ln\pi_\theta(a|s)(R_t-V_t(s_t))}$.
Here we can train the critic to approximate $V(s_t)$ using familiar temporal
difference methods (Section~\ref{bootstrapping}). This algorithm
is known as \algo{AdvantageActorCritic} as $R_t-V_t(s_t)$ estimates the
advantage function $Q(s_t,a_t)-V(s_t)$ that describes how good an action is
compared to how good the average action is in the current state.

\chapter{Challenges in Complex Environments}
\label{challenges}

Traditional benchmark problems in RL include tasks like \task{Mountain Car} and
\task{Cart Pole}. In those tasks, the observation and action spaces are small,
and the dynamics can be described by simple formulas. Moreover, these
environments are usually fully observed so that the agent could reconstruct the
system dynamics perfectly, given enough transitions.

In contrast, 3D environments like \env{Doom} have complex underlying dynamics
and large state spaces that the agent can only partially observe. Therefore, we
need more advanced methods to learn successful policies
(Chapter~\ref{algorithms}). We now discuss challenges that arise in complex
environments and methods to approach them.

\section{Large State Space}
\label{large_state_space}


Doom is a 3D environment where agents observe perspective 2D projections from
their position in the world as pixel matrices. Having such a large state space
makes tabular versions of RL algorithms intractable. We can adjust those
algorithms to the use of function approximators and make them tractable.
However, the agent receives tens of thousands of pixels every time step. This
is a computational challenge even in the case of function approximation.

Downsampling input images only provides a partial solution to this since we
must preserve information necessary for effective control. We would like our
agents to find small representations of its observations that are helpful for
action selection. Therefore, abstraction from individual pixels is necessary.
\emph{Convolutional neural networks} (CNNs) provide a computationally effective
way to learn such abstractions~\cite{mnih2013dqn}.

In comparison to normal \emph{fully-connected neural networks}, CNNs consist of
\emph{convolutional layers} that learn multiple filters. Each filter is shifted
over the whole input or previous layer to produce a feature map. Feature maps
can optionally be followed by \emph{pooling layers} that downsample each
feature map individually, using the max or mean of neighboring pixels.

Applying the filters across the whole input or previous layer means that we
must only learn a small filter kernel. The number of parameters of this kernel
does not depend on the input size. This allows for more efficient computation
and faster learning compared to fully-connected networks where a layers adds an
amount of parameters that is quadratic in the number of the layer size.

Moreover, CNNs exploit the fact that nearby pixels in the observed images are
correlated. For example, walls and other surfaces result in evenly textured
regions. Pooling layers help to reduce the dimensionality while keeping
information represented small details, when important. This is because each
filter learns a high-resolution feature and is downsampled individually.

\section{Partial Observability}
\label{partial_observability}

In complex environments, observations do no fully reveal the state of the
environment. The perspective view that the agent observes in the Doom
environment contains reduced information in multiple ways:

\begin{itemize}
\item The agent can only look in forward direction and its field of view only
includes a fraction of the whole environment.
\item Obstacles like walls hide the parts of the scene behind them.
\item The perspective projection loses information about the 3D geometry of the
scene. Reconstructing the 3D geometry and thus positions of objects is not
trivial and might not even have a unique solution.
\item Many 3D environments include basic physics simulations. While the agent
can see the objects in its view, the pixels do not directly contain
information about velocities. It might be possible to infer them, however.
\item Several other temporal factors are not represented in the current image,
like whether an item or enemy in another room still exists.
\end{itemize}

To learn a good policy, the agent has to detect spatial and temporal
correlations in its input. For example, it would be beneficial to know the
position of objects and the agent itself in the 3D environment. Knowing the
positions and velocities of enemies would certainly help aiming.

Using hierarchical function approximators like neural networks allows learning
high-level representations like the existence of an enemy in the field of view.
For high-level representations, neural networks need more than one layer
because a layer can only learn linear combinations of the input or previous
layer. Complex features might be impossible to construct from a linear
combination of input pixels. \citet{zeiler2014visualizing} visualize the layers
of CNNs and show that they actually learn more abstract features in each layer.

To address the temporal incompleteness of the observations, we can use
\emph{frame skipping}, where we collect multiple images and show this stack as
one observation to the agent. The agent then decides for an action we repeat
while collecting the next stack of inputs~\cite{mnih2013dqn}.

It is also common to use \emph{recurrent neural networks}
(RNNs)~\cite{hausknecht2015drqn,mnih2016a3c} to address the problem of partial
observability. Neurons in these architectures have self-connections that allow
activations to span multiple time steps. The output of an RNN is a function of
the current input and the previous state of the network itself. In particular,
a variant called \emph{Long Short-Term Memory} (LSTM) and its variations like
\emph{Gated Recurrent Unit} (GRU) have proven to be effective in a wide range
of sequential problems~\cite{lipton2015rnn}.

\begin{figure}
\centering
\includegraphics[width=\textwidth]{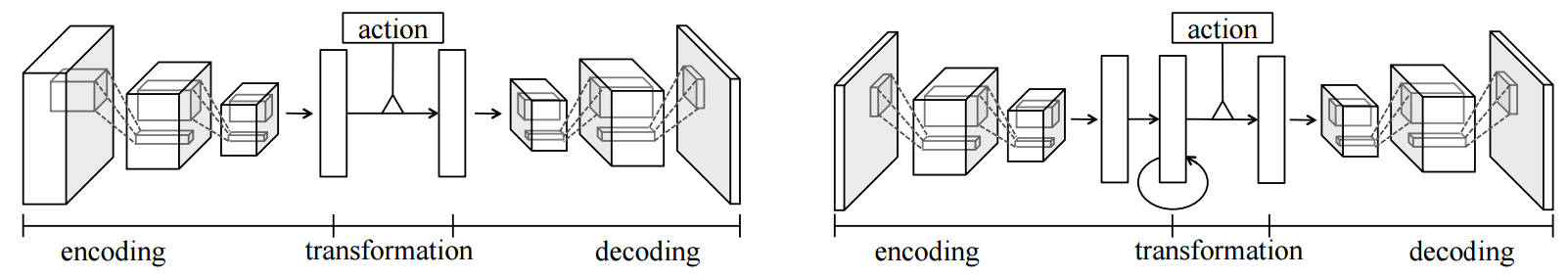}
\caption{Two neural network architectures for learning representations in POMDPs
that were used for predicting future observations in \env{Atari}. (From
\citet{oh2015videoprediction})}
\label{fig:video_prediction}
\end{figure}

Combining CNNs and LSTMs, \citet{oh2015videoprediction} were able to learn
useful representations from videos, allowing them to predict up to $100$
observations in the \env{Atari} domain (Figure~\ref{fig:video_prediction}).

A recent advancement was applying \emph{memory network}
architectures~\cite{weston2014memory,graves2014ntm}\ to RL problems in the
Minecraft environment~\cite{oh2016frmqn}. These approximators consist of an RNN
that can write to and read from an external memory. This allows the network to
clearly carry information over long durations.

\section{Stable Function Approximation}

Various forms of neural networks have been successfully applied to supervised
and unsupervised learning problems. In those applications, the dataset is often
known prior to training and can be decorrelated and many machine learning
algorithms expect \emph{independent and identically distributed} data. This is
problematic in the RL setting, where we want to improve the policy while
collecting observations sequentially. The observations can be highly correlated
due to the sequential nature underlying the MDP.

To decorrelate data, the agent can use a \emph{replay memory} as introduced by
\citet{mnih2013dqn} to store previously encountered transitions. At each time
step, the agent then samples a random batch of transitions from this memory and
uses this for training. To initialize the memory, one can run a random policy
before the training phase. Note that the transitions are still biased by the
start distribution of the MDP.

The mentioned work first managed to learn to play several 2D Atari
games~\cite{bellemare13ale} without the use of hand-crafted features. It has
also been applied to simple tasks in the
\env{Minecraft}~\cite{trevor2016blockworld} and
\env{Doom}~\cite{kempka2016vizdoom} domains. On the other hand, the replay
memory is memory intense and can be seen as unsatisfyingly far from the way
humans process information.

In addition, \citet{mnih2013dqn} used the idea of a \emph{target network}. When
training approximators by bootstrapping (Section~\ref{bootstrapping}), the
targets are estimated by the same function approximator. Thus, after each
training step, the target computation changes which can prevent convergence as
the approximator is trained toward a moving target. We can keep a previous
version of the approximator to obtain the targets. After every few time steps,
we update the target network by the current version of the training
approximator.

Recently, \citet{mnih2016a3c} proposed an alternative to using replay memories
that involves multiple versions of the agent simultaneously interacting with
copies of the environment. The agents apply gradient updates to a shared set of
parameters. Collecting data from multiple environments at the same time
sufficiently decorrelated data to learn better policies than replay memory and
target network were able to find.

\section{Sparse and Delayed Rewards}

The problem of non-informative feedback is not tied to 3D environments in
particular, yet constitutes a significant challenge. Sometimes, we can help and
guide the agent by rewarding all actions with positive or negative rewards. But
in many real-world tasks, the agent might receive zero rewards most of the time
and only see a binary feedback at the end of each episode.

Rare feedback is common when hand-crafting a more informative reward signal is
not straightforward, or when we do not want to bias the solutions that the
agent might find. In these environments, the agent has to assign credit among
all its previous actions when finally receiving a reward.

We can apply eligibility traces (Section~\ref{eligibility_traces}) to function
approximation by keeping track of all transitions since the start of the
episode. When the agent receives a reward, it incorporates updates for all
stored states relative to the decayed reward.


Another way to address sparse and delayed rewards is to employ methods of
temporal abstraction as explained in Section~\ref{temporal_abstraction}

\section{Efficient Exploration}
\label{exploration}

Algorithms like \algo{Q-Learning} (Section~\ref{bootstrapping}) are optimal in
the tabular case under the assumption that each state will be visited over and
over again, eventually~\cite{tsitsiklis1994qoptimal}. In complex environments,
it is impossible to visit each of the many states. Since we use function
approximation, the agent can already generalize between similar states. There
are several paradigms to the problem of effectively finding interesting and
unknown experiences that help improve the policy.

\subsection{Random Exploration}

We can use a simple \algo{EpsilonGreedy} strategy for exploration, where at
each time step, with a probability of $\varepsilon\in \left(0,1\right]$, we
choose a completely random action, and act according to our policy otherwise.
When we start at $\varepsilon=1$ and exponentially decay $\varepsilon$ over
time, this strategy, in the limit, guarantees to visit each state over and over
again.

In simple environments, \algo{EpsilonGreedy} might actually visit each state
often enough to derive the optimal policy. But in complex 3D environments, we
do not even visit each state once in a reasonable time. Visiting novel states
would be important to discover better policies.

One reason that random exploration still works reasonably well in complex
environments~\cite{mnih2013dqn,mnih2016a3c} can partly be attributed to
function approximation. When the function approximator generalizes over similar
states, visiting one state also improves the estimate of similar states.
However, more elaborate methods exist and can result in better results.

\subsection{Optimism in the Face of Uncertainty}
\label{uncertainty}

A simple paradigm to encourage exploration in value-based algorithms
(Section~\ref{value_based}) is to optimistically initialize the estimated
Q-values. We can do this either by pre-training the function approximator or by
adding a positive bias to its outputs. Whenever the agent visits an unknown
state, it will correct its Q-value downward. Less visited states still have
high values assigned so that the agent tends to visiting them when facing the
choice.

The Bayesian approach is to count the visits of each state to compute the
uncertainty of its value estimate~\cite{strens2000bayesian}. Combined with
Q-learning, this converges to the true Q-function, given enough random
samples~\cite{li2009theory}. Unfortunately, it is hard to obtain truly random
samples for a general MDP because of its sequential nature. Another problem of
counting is that the state space may be large or continuous and we want to
generalize between similar states.

\citet{bellemare2016countdensity} recently suggested a sequential density
estimation model to derive pseudo-counts for each state in a non-tabular
setting. Their method significantly outperforms existing methods on some games
of the \env{Atari} domain, where exploration is challenging.

We can also use uncertainty-based exploration with policy gradient methods
(Section~\ref{policy_based}). While we usually do not estimate the Q-function
here, we can add the entropy of the policy as a \emph{regularization term} to
the its gradient~\cite{williams1991entropy} with a small factor $\beta\in
\mathbb{R}$ specifying the amount of regularization:

\begin{equation}
\label{entropy_regularization}
\begin{gathered}
\E{\pi_\theta}{R(s,a)\nabla_\theta\log\pi_\theta(a|s)}
+\beta\nabla_\theta \mathbb{H}\big(\pi_\theta(a|s)\big), with\\
\mathbb{H}\big(\pi_\theta(a|s)\big)=
-\sum_{a'\in A}{\pi_\theta(a'|s)}\log\pi_\theta(a'|s).
\end{gathered}
\end{equation}

This encourages a uniformly distributed policy until the policy gradient
updates outweigh the regularization. The method was successfully applied to
\env{Atari} and a visual 3D domain \env{Labyrinth} by \citet{mnih2016a3c}.

\subsection{Curiosity and Intrinsic Motivation}

A related paradigm is to explore in order to attain knowledge about the
environment in the absence of external rewards. This usually is a
\emph{model-based} approach that directly encourages novel states based on two
function approximators.

The first approximator, called \emph{model}, tries to predict the next
observation and thus approximates the transition function of the MDP. The
second approximator, called \emph{controller}, performs control. Its objective
is both to maximize expected returns and to cause the highest reduction in
prediction error of the model~\cite{jurgen1991modelcontroller}. It therefore
tries to provide new observations to the model that are novel but learnable,
inspired by the way humans are bored by both known knowledge and knowledge they
cannot understand~\cite{schmidhuber2010creativity}.

The model-controller architecture has been extended in multiple ways.
\citet{ngo2013selfgoals} combined it with planning to escape known areas of the
state space more effectively. \citet{schmidhuber2015think} recently proposed
shared neurons between the model and controller networks in a way that allows
the controller to arbitrarily exploit the model for control.

\subsection{Temporal Abstraction}
\label{temporal_abstraction}

While most RL algorithms directly produce an action at each time step, humans
plan actions on a slower time scale. Adapting this property might be necessary
for human level control in complex environments. Most of the mentioned
exploration methods (Section~\ref{exploration}) determine the next exploration
action at each time step. Temporally extended actions could be beneficial to
both exploration and exploitation~\cite{rasmussen2014hierarchic}.

The most common framework for temporal abstraction in the RL literature is the
\emph{options framework} proposed by \citet{sutton1999options}. The idea is to
learn multiple low-level policies, named \emph{options}, that interact with the
world. A high-level policy observes the same inputs but has the options as
actions to choose from. When the high-level policy decides for an option, the
corresponding low-level policy is executed for a fixed or random number of time
steps.

While there are multiple ways to obtain options, two recent approaches were
shown to work in complex environments. \citet{krishnamurthy2016spectral} used
spectral clustering to group states with cluster centers representing options.
Instead of learning individual low-level policies, the agent greedily follows a
distance measure between low-level states and cluster centers that is given by
the clustering algorithm.

Also building on the options framework, \citet{tessler2016dsn} trained multiple
CNNs on simple tasks in the \env{Minecraft} domain. These so-called \emph{deep
skill networks} are added in addition to the low-level actions for the
high-level policy to choose from. The authors report promising results on a
navigation task.

A limitation of the options framework is its single level of hierarchy. More
realistically, multi-level hierarchical algorithms are mainly explored in the
fields of cognitive science and computational neuroscience.
\citet{rasmussen2014hierarchic} propose one such architecture and show that it
is able to learn simple visual tasks.

\chapter{Algorithms for Learning from Pixels}
\label{algorithms}

We explained the background of RL methods in Chapter~\ref{background} and
described the challenges that arise in complex environments in
Chapter~\ref{challenges}, where we already outlined the intuition behind some
of the current algorithms. In this section, we build on this and explain two
state-of-the-art algorithms that have successfully been applied to 3D domains.

\section{Deep Q-Network}
\label{dqn}

The currently most common algorithm for learning in high-dimension state spaces
is the \emph{Deep Q-Network} (\algo{DQN}) algorithm suggested by
\citet{mnih2013dqn}. It is based on traditional \algo{Q-Learning} algorithm
with function approximation and \algo{EpsilonGreedy} exploration. In its
initial form, \algo{DQN} does not use eligibility traces.


The algorithm uses a two-layer CNN, followed by a linear fully-connected layer
to approximate the Q-function. Instead of taking both state and action as
input, it outputs the approximated Q-values for all actions $a\in A$
simultaneously, taking only a state as input.


To decorrelate transitions that the agent collects during play, it uses a large
replay memory. After each time step, we select a batch of transitions from the
replay memory randomly. We use the temporal difference \algo{Q-Learning} rule to
update the neural network. We initialize $\varepsilon=1$ and start annealing it
after the replay memory is filled.

In order to further stabilize training, \algo{DQN} uses a target network to
compute the temporal difference targets. We copy the weights of the primary
network to the target network at initialization and after every few time steps.
As initially proposed, the algorithms synchronizes the target network every
frame, so that the targets are computed using the network weights at the last
time step.


\algo{DQN} has been successfully applied to two tasks within the \env{Doom}
domain by \citet{kempka2016vizdoom} who introduced this domain. In the more
challenging \env{Health Gathering} task, where the agent must collect health
items in multiple open rooms, they used three convolutional layers, followed by
max-pooling layers and a fully-connected layer. With a small learning rate of
$\alpha=0.00001$, a replay memory of $10000$ entries, and one million training
steps, the agent learned a successful policy.

\citet{trevor2016blockworld} applied \algo{DQN} to two tasks in the
\env{Minecraft} domain: Collecting as many blocks of a certain color as
possible, and navigating forward on a pathway without falling. In both tasks,
\algo{DQN} learned successful policies. Depending on the width of the pathway,
a larger convolutional network and several days of training were needed to
learn successful policies.

\section{Asynchronous Advantage Actor Critic}
\label{a3c}

\algo{A3C} by \citet{mnih2016a3c} is an actor-critic method that is
considerable more memory-efficient than \algo{DQN}, because it does not require
the use of a replay memory. Instead, transitions are decorrelated by training
in multiple versions of the same environment in parallel and asynchronously
updating a shared actor-critic model. Entropy regularization
(Section~\ref{uncertainty}) is employed to encourage exploration.

Each of the originally up to $16$ threads manages a copy of the model, and
interacts with an instance of the environment. Each threads collects a few
transitions before computing eligibility returns
(Section~\ref{eligibility_traces}) and computing gradients according to the
\algo{AAC} algorithm (Section~\ref{actor_critic}) based on its current copy of
the model. It then applies this gradient to the shared model and updates its
copy to the current version of the shared model.


One version of \algo{A3C} uses the same network architecture as \algo{DQN},
except for using a softmax activation function in the last layer to model the
policy rather than. The critic model shares all convolutional layers and only
adds a linear fully-connected layer of size one that is trained to estimate the
value function. The authors also proposed a version named \algo{LSTM-A3C} that
adds one LSTM layer between the convolutional layers and the output layers to
approach the problem of partial observability.

In addition to \env{Atari} and a continuous control domain, \algo{A3C} was
evaluated on a new \env{Labyrinth} domain, a 3D maze where the goal is to find
and collect as many apples as possible. \algo{LSTM-A3C} was able to learn a
successful policy for this task that avoids to walk into walls and turns around
when facing dead ends. Given the recent proposal of the algorithm, it is not
likely that the algorithm was applied to other 3D domains yet.

%
%
%
%
%

\chapter{Experiments in the Doom Domain}
\label{experiments}

We now introduce the \env{Doom} domain in detail, focusing on the task used in
the experiments. We explain methods that were necessary to train the algorithms
(Chapter~\ref{algorithms}) in a stable manner. While the agents do not reach
particularly high scores on average, they learn a range of interesting
behaviors that we examine in Section~\ref{results}.

\section{The Doom Domain}
\label{doom_domain}

The \env{Doom} domain~\cite{kempka2016vizdoom} provides RL tasks simulated by
the game mechanics of the first-person shooter Doom. This 3D game features
different kinds of enemies, items, and weapons. A level editor can be used to
create custom tasks. We use the \task{DeathMatch} task defined in the Gym
collection~\cite{brockman2016gym}. We first describe the \env{Doom} domain in
general.

In \env{Doom}, the agent observes image frames that are perspective 2D
projections of the world from the agent's position. A frame also contains user
interface elements at the bottom, including the amount of ammunition of the
agent's selected weapon, the remaining health points of the agent, and
additional game-specific information. We do not extract this information
explicitly.

Each frame is represented as a tensor of dimensions \emph{screen width},
\emph{screen height}, and \emph{color channel}. We can choose the width and
height from a set of available screen resolutions. The color channel represents
the RGB values of the pixels and thus always has a size of three.

The actions are arbitrary combinations of the $43$ available user inputs to the
game. Most actions are binary values to represent whether a given key on the
keyboard is in pressed or released position. Some actions represent mouse
movement and take on values in the range $\left[-10,10\right]$ with $0$ meaning
no movement.

The available actions perform commands in the game, such as \emph{attack},
\emph{jump}, \emph{crouch}, \emph{reload}, \emph{run}, \emph{move left},
\emph{look up}, \emph{select next weapon}, and more. For a detailed list,
please refer to the whitepaper by \citet{kempka2016vizdoom}.

We focus on the \task{DeathMatch} task, where the agent faces multiple kinds of
enemies that attack it. The agent must shoot an enemy multiple times, depending
on the kind of enemy, to kill it an receive a reward of $+1$. Being hit by
enemy attacks reduces the agent's health points, eventually causing the end of
the episode. The agent does not receive any reward the end of the episode. The
episode also ends after exceeding $10^4$ time steps.

\begin{figure}
\centering
\includegraphics[width=0.6\textwidth]{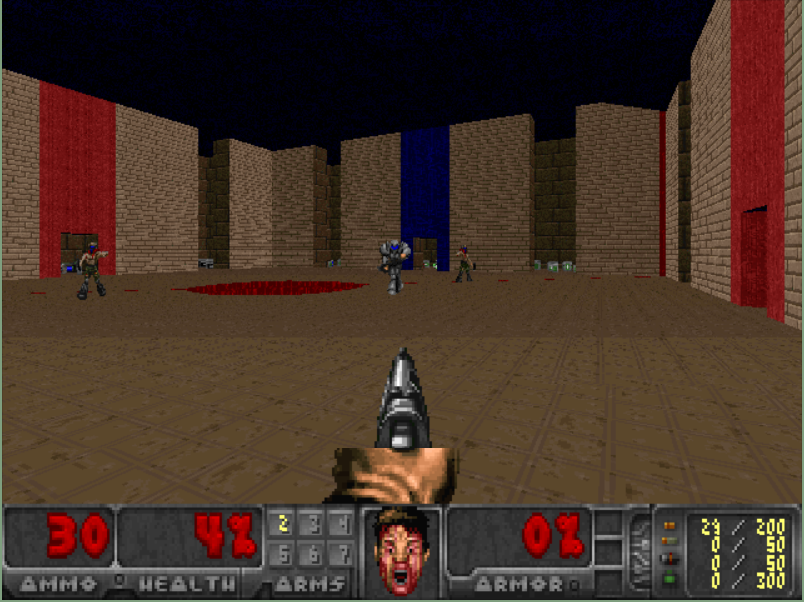}
\caption{The environment of the \task{DeathMatch} task contains a hall where
enemies appear, and four rooms to the sides. The rooms contain either health
items (\emph{red entry}) or weapons and ammunition (\emph{blue entry}).}
\label{fig:doom_map}
\end{figure}

As shown in Figure~\ref{fig:doom_map}, the world consists of one large hall,
where enemies appear regularly, and four small rooms to the sides. Two of the
small rooms contain items for the agent to restore health points. The other two
rooms contain ammunition and stronger weapons than the pistol that the agent
begins with. To collect items, ammunition, or weapons, the agent must walk
through them. The agent starts at a random position in the hall, facing a
random direction.

\section{Applied Preprocessing}
\label{preprocessing}

\begin{figure}
\centering
\includegraphics[width=0.3\textwidth]{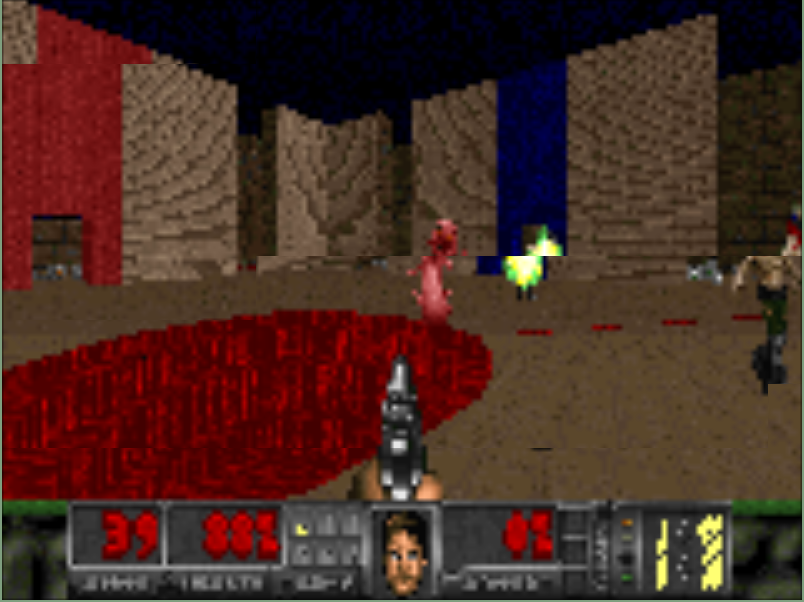}
\includegraphics[width=0.3\textwidth]{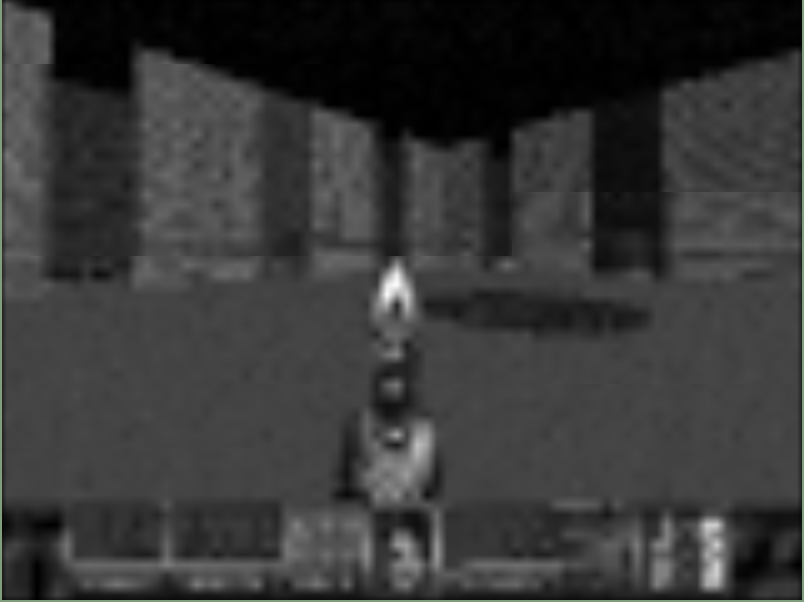}
\includegraphics[width=0.3\textwidth]{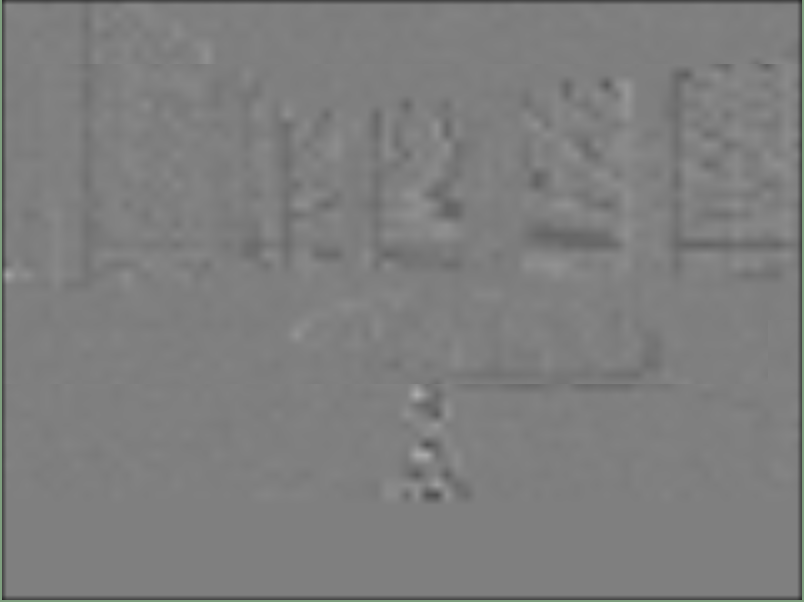}
\caption{Three observations of the \task{DeathMatch} task: An unprocessed frame
(\emph{left}), a downsampled grayscale frame (\emph{center}), and a downsampled
grayscale delta frame (\emph{right}). Based on human testing, both
preprocessing methods retain observations sufficient for successful play.}
\label{fig:doom_observations}
\end{figure}

To reduce computation requirements, we choose the smallest available screen
resolution of $160\times 120$ pixels. We further down-sample the observations
to $80\times 60$ pixels, and average over the color channels to produce a
grayscale image. We experiment with \emph{delta frames}, where we pass the
difference between the current and the last observation to the agent. Both
variants are visualized in Figure~\ref{fig:doom_observations}.

We further employ \emph{history frames} as originally used by \algo{DQN} in the
\env{Atari} domain, and further explored by \citet{kempka2016vizdoom}. Namely,
we collect multiple frames, stack them, and show them to the agent as one
observation. We then repeat the agent's action choice over the next time steps
while collecting a new stack of images. We perform a random action during the
first stack of an episode. History frames have multiple benefits: They include
temporal information, allow more efficient data processing, and cause actions
to be extended for multiple time steps, resulting in more smooth behavior.

History frames extend the dimensionality of the observations. We use the
grayscale images to compensate for this and keep computation requirements
manageable. While color information could be beneficial to the agent, we expect
the temporal information contained in history frames to be more valuable,
especially to the state-less \algo{DQN} algorithm. Further experiments would be
needed to test this hypothesis.

We remove unnecessary actions from the action space to speed up learning,
leaving only the $7$ actions \emph{attack}, \emph{move left}, \emph{move
right}, \emph{move forward}, \emph{move backward}, \emph{turn left}, and
\emph{turn right}. Note that we do not include actions to rotate the view
upward or downward, so that the agent does not have to learn to look upright.

Finally, we apply normalization: We scale the observed pixels into the range
$\left[0,1\right]$ and normalize rewards to $(-1,0,+1)$ using
$r_t\leftarrow\sgn(r_t)$~\cite{mnih2013dqn}. Further discussion of
normalization can be found in the next section.

\section{Methods to Stabilize Learning}

Both \algo{DQN} and \algo{LSTM-A3C} are sensitive to the choice of hyper
parameters~\cite{mnih2016a3c,sprague2015sensitivity}. Because training times
lie in the order of hours or days, it is not tractable to perform an excessive
hyper parameter search for most researcher. We can normalize observations and
rewards as described in the previous section to make it more likely that hyper
parameters can be transferred between tasks and domains.

It was found to be essential to clip gradients of the networks in both
\algo{DQN} and \algo{LSTM-A3C}. Without this step, the approximators diverges
in the early phase of learning. Specifically, we set each element $x$ of the
gradient to $x\leftarrow\max\left\{-10,\min\left\{x,+10\right\}\right\}$. The
exact threshold to clip at did not seem to have a large impact on training
stability or results.

We decay the learning rate linearly over the course of training to guarantee
convergence, as done by \citet{mnih2013dqn,mnih2016a3c}, but not by
\citet{kempka2016vizdoom,trevor2016blockworld}.

\section{Evaluation Methodology}

\algo{DQN} uses a replay memory of the last $10^4$ transitions and samples
batches of size $64$ from it, following the choices by
\citet{kempka2016vizdoom}. We anneal $\varepsilon$ from $1.0$ to $0.1$ over the
course of $2*10^6$ time steps, which is one third of the training duration. We
start both training and annealing $\varepsilon$ after the first $10^4$
observations. This is equivalent to initializing the replay memory from
transitions collected by a random policy.

For \algo{LSTM-A3C}, we use $16$ learning threads that apply their accumulated
gradients every $5$ observations using shared optimizer statistics, a entropy
regularization factor of $\beta=0.01$. Further, we scale the loss of the critic
by~$0.5$~\cite{mnih2016a3c}.

Both algorithms operate on frames of history $6$ at a time and use RMSProp with
a decay parameter of $0.99$ for optimization~\cite{mnih2016a3c}. The learning
rate starts at $\alpha=2*10^{-5}$, similar to \citet{kempka2016vizdoom} who use
a fixed learning rate of $10^{-5}$ in the \env{Doom} domain.

We train both algorithms for $20$ epochs, where one epoch corresponds to
$5*10^4$ observations for \algo{DQN} and $8*10^5$ observations for
\algo{LSTM-A3C}, resulting in comparable running times of both algorithms.
After each epoch, we evaluate the learned policies on $10^4$ or $10^5$
observations, respectively. We measure the score, that is, the sum of collected
rewards, that we average over the training episodes. \algo{DQN} uses
$\varepsilon=0.05$ during evaluation.

\section{Characteristics of Learned Policies}
\label{results}

In the conducted experiments, both algorithms were able to learn interesting
policies during the $20$ epochs. We describe and discuss instances of such
policies in this section. The results of \algo{DQN} and \algo{LSTM-A3C} are
similar in many cases, suggesting that both algorithms discover common
structures of the \task{DeathMatch} task.

The agents did not show particular changes when receiving delta frames
(Section~\ref{preprocessing}) as inputs. This might be because given history
frames, neural networks can easily learn to compute differences between those
frames if useful to solve the task. Therefore, we conclude that delta frames
may not be useful in combination with history frames.

\subsection{Fighting Against Enemies}

As expected, the agents develop a tendency to aim at enemies. Because an agent
needs to directly look toward enemies in order to shoot them, it always faces
an enemy the time step before receiving a positive reward. However, the aiming
is inaccurate and the agents tend to look past their enemies to the left and
right side, alternatingly. It would be interesting to conduct experiments
without stacking history frames to understand whether the learned policies are
limited by the action repeat.

In several instances, the agents lose enemies from their field of view by
turning, usually toward other enemies. No trained agent was found to memorize
such enemies after not seeming them anymore. As a result, the agents were often
attacked from the back, causing the episode to end.

Surprisingly, \algo{LSTM-A3C} agents tend to only attack when facing an enemy,
but sometimes miss out the chance to do so. In contrast, \algo{DQN} agents were
found to shoot regularly and more often than \algo{LSTM-A3C}. Further
experiments would be needed to verify, whether this behavior persists after
training for more than $20$ episodes.

\subsection{Navigation in the World}

Most agents learn to avoid running into walls while facing them, suggesting at
least a limited amount of spatial awareness. In comparison, a random agent
repeatedly runs into a wall and gets stuck in this position until the end of
the episode.

In some instances, the agent walks backward against a wall at an angle, so that
it slides along the wall until reaching the entrance of a side room. This
behavior overfits to the particular environment of the \task{DeathMatch} task,
but represents a reasonable policy to pick up the more powerful weapons inside
two of the side rooms. We did not find testing episodes in which the agent used
this strategy to enter one of the rooms containing health restoring items. This
could be attributed to the lack of any negative reward when the agent dies.

Notably, in one experiment, \algo{LSTM-A3C} agent collects the rocket launcher
weapon using this wall-sliding strategy, and starts to fire after collecting
it. The agents did not fire before collecting that weapon. During training,
there might have been successful episodes, where the agent initially collected
this weapon. Another reason for this could be that the first enemies only
appear after a couple of time steps so there is no reason to fire in the
beginning of an episode.

In general, both tested algorithms, \algo{DQN} and \algo{LSTM-A3C}, were able
to learn interesting policies for our task in the \env{Doom} domain in
relatively short amounts of training time. On the other hand, the agents did
not find ways to effectively shoot multiple enemies in a row. An evaluation
over multiple choices of hyper parameters and longer training times would be a
sensible next step.

\chapter{Conclusions}
\label{conclusions}

We started by introducing reinforcement learning (RL) and summarized basic
reinforcement learning methods. Compared to traditional benchmarks, complex 3D
environments can be more challenging in many ways. We analyze these challenges
and review current algorithms and ideas to approach these.

Following this background, we select to algorithms that were successfully
applied to learn from pixels in 3D environments. We apply both algorithms to a
challenging task within the recently presented Doom domain, that is based on a
3D first-person-shooter game.

Applying the selected algorithms yields policies that do not reach particularly
high scores, but show interesting behavior. The agents tend to aim at enemies,
avoid getting stuck in walls, and find surprising strategies within this Doom
task.

We also assess that using delta frames does not affect the training much when
combined with stacking multiple images together as input to the agent. In would
be interesting to see if using fewer action repeats allows for more accurate
aiming.

The drawn conclusions are not final because other hyper parameters or longer
training might allow these algorithms to perform better. We analyze instances
of learned behavior and believe that observations reported in the work provide
a valuable starting point for further experiments in the \env{Doom} domain.


\addcontentsline{toc}{chapter}{Bibliography}
\bibliography{references}

\chapter*{Declaration of Authorship}

\thispagestyle{empty}

I hereby declare that the thesis submitted is my own unaided work. All direct
or indirect sources used are acknowledged as references.\vspace{2 ex}

Potsdam, \today\\[6 ex]

\begin{flushleft}
    \begin{tabular}{p{5cm}}
        \hline
        \centering\footnotesize Danijar Hafner
    \end{tabular}
\end{flushleft}

\end{document}